\documentclass{article}
\usepackage[accepted]{icml2024}

\usepackage{natbib}
\setcitestyle{authoryear}
\usepackage{caption}
\usepackage{hyperref}
\usepackage{breakurl}
\usepackage{tightenum}
\usepackage{wrapfig}
\usepackage{pifont}
\usepackage{soul}
\usepackage{enumitem}
\usepackage{listings}
\usepackage{latexsym}
\usepackage{adjustbox}
\usepackage{graphicx}
\usepackage{floatflt}
\usepackage{setspace}
\usepackage{graphicx}
\usepackage{algorithm}
\usepackage{amsmath}
\usepackage{cleveref}
\usepackage[small,compact]{titlesec}
\usepackage{titling}
\usepackage{authblk}
\usepackage{xspace}
\usepackage{makecell}
\usepackage{caption}

\hypersetup{
    colorlinks,
    linkcolor={red!50!black},
    citecolor={blue!50!black},
    urlcolor={blue!80!black}
}

\usepackage[normalem]{ulem}

\newcommand{\us}[1]{$\mu$s}
\newcommand{\fixme}[1]{{\color{red}\textbf{#1}}}
\newcommand{\x}[1]{$\times$}
\definecolor{pink}{rgb}{1.0,0.47,0.6}
\definecolor{orange}{rgb}{1.0,0.5,0.0}
\definecolor{cyanish}{rgb}{0,0.8,1.0}

\newcommand{\mycommand}[1]{\texttt{#1}}
\newcommand{\LogWrite}[1]{\mycommand{LogWrite}}
\newcommand{\Commit}[1]{\mycommand{Commit}}
\newcommand{\Abort}[1]{\mycommand{Abort}}
\newcommand{\WriteBack}[1]{\mycommand{WriteBack}}
\newcommand{\AtomicWrite}[1]{\mycommand{AtomicWrite}}
\newcommand{\ShortAtomicWrite}[1]{\mycommand{ShortAtomicWrite}}

\newcommand{\mystate}[1]{\textsc{#1}}
\newcommand{\Free}[1]{\mystate{Free}}
\newcommand{\Pending}[1]{\mystate{Pending}}
\newcommand{\Committed}[1]{\mystate{Committed}}

\newcommand{\Valid}[1]{\mystate{Valid}}
\newcommand{\Marked}[1]{\mystate{Marked}}

\newcommand{\eg}{\textit{e.g.}}
\newcommand{\ie}{\textit{i.e.}}

\newcommand{\beforecaption}{\vspace{-.15cm}\begin{spacing}{0.85}}
\newcommand{\aftercaption}{\vspace{-.45cm}\end{spacing}}
\newcommand{\mycaption}[3]{\beforecaption\caption{\label{#1}{\bf \footnotesize #2} \em\footnotesize #3}\aftercaption}

\newcommand{\trick}[0]{\textsc{InferCept}\xspace}
\newcommand{\sys}{\textsc{InferCept}\xspace}

\newcommand{\discard}{\texttt{Discard}\xspace}
\newcommand{\improvediscard}{\texttt{ImprovedDiscard}\xspace}
\newcommand{\preserve}{\texttt{Preserve}\xspace}
\newcommand{\swap}{\texttt{Swap}\xspace}

\newcommand{\yiying}[1]  {\noindent{\color{blue} {\bf \fbox{Yiying}     {\it#1}}}}
\newcommand{\reyna}[1]  {\noindent{\color{orange} {\bf \fbox{Reyna}     {\it#1}}}}

\newcommand{\zijian}[1]  {\noindent{\color{orange} {\bf \fbox{Zijian}     {\it#1}}}}

\usepackage{calc}

\usepackage[textsize=tiny,textwidth=0.6in]{todonotes}
\newcommand{\allnotes}[1]{}

\renewcommand{\em}{\it}

\newcommand{\ignore}[1]{}

\newcommand{\boldunderpara}[1]{\noindent{\underline{\textbf{#1}}}}

\def\cffigure[#1,#2,#3]{
\begin{floatingfigure}{3.5in}
\vspace*{-2mm}
\begin{center}

\includegraphics[width=3in]{#1} 
 
\vspace*{-3mm}\caption[]{#2\vspace*{3ex}}
\label{#3} 
\vspace*{-5mm}
\end{center}
\vspace*{-2mm}
\end{floatingfigure}}

\def\cfigure[#1,#2,#3]{
\begin{figure}
\vspace*{0mm}
\begin{center}

\includegraphics[width=3in]{#1} 
 
\vspace*{-3mm}\caption[]{#2
} \label{#3}
 
\vspace*{-5mm}
\end{center}
\vspace*{-2mm}
\end{figure}}

\def\wfigure[#1,#2,#3]{
\begin{figure*}
\vspace*{0mm}
\begin{center}
 
\includegraphics[width=6in]{#1} 
 
\vspace*{-3mm}\caption[]{#2
} \label{#3}
 
\vspace*{-5mm}
\end{center}
\vspace*{-2mm}
\end{figure*}}

\def\dcfigure[#1,#2,#3,#4,#5,#6]{
{
\begin{figure*}
\vspace*{0.0in}\
\begin{center}
\begin{minipage}[c]{3in}{
\includegraphics[width=3in]{#1} 
\vspace*{-3mm}\caption[]{#2} \label{#3} \
}\end{minipage}\hspace*{0.5in}\
\begin{minipage}[c]{3in}{
\includegraphics[width=3in]{#4} 
\vspace*{-3mm}\caption[]{#5}\label{#6} \
}\end{minipage}
\end{center}
\vspace*{-0.4in}\
\end{figure*}
}
}

\def\threefigure[#1,#2,#3,#4,#5]{
\begin{figure*}
\vspace*{0mm}
\begin{center}

\begin{tabular}{ccc}
\includegraphics[width=2in]{#1} & \includegraphics[width=2in]{#2} &  \includegraphics[width=2in]{#3} \\
(a) & (b) & (c) \\
\end{tabular}

\vspace*{-3mm}\caption[]{#4
} \label{#5}

\vspace*{-5mm}
\end{center}
\vspace*{-2mm}
\end{figure*}}

\def\dssfigure[#1,#2,#3,#4,#5,#6]{
{
\begin{figure*}
\vspace*{0.2in}\
\begin{center}
\begin{minipage}[c]{4in}{
\includegraphics[width=4in]{#1}
\vspace*{-3mm}\caption[]{#2} \label{#3} \
}\end{minipage}\hspace*{0.5in}\
\begin{minipage}[c]{2in}{
\includegraphics[width=2in]{#4}
\vspace*{-3mm}\caption[]{#5}\label{#6} \
}\end{minipage}
\end{center}
\vspace*{-0.4in}\
\end{figure*}
}
}

\def\dsfigure[#1,#2,#3,#4,#5,#6]{
{
\begin{figure*}
\vspace*{0.2in}\
\begin{center}
\begin{minipage}[c]{3in}{
\includegraphics[width=3in]{#1}
\vspace*{-3mm}\caption[]{#2} \label{#3} \
}\end{minipage}\hspace*{0.5in}\
\begin{minipage}[c]{3in}{
\hspace*{0.5in}\
\includegraphics[height=3in]{#4}
\vspace*{-3mm}\caption[]{#5}\label{#6} \
}\end{minipage}
\end{center}
\vspace*{-0.4in}\
\end{figure*}
}
}

\def\dsyfigure[#1,#2,#3,#4,#5,#6]{
{
\begin{figure*}
\vspace*{0.2in}\
\begin{center}
\begin{minipage}[c]{2.5in}{
\includegraphics[height=2.5in]{#1}
\vspace*{-3mm}\caption[]{#2} \label{#3} \
}\end{minipage}\hspace*{0.5in}\
\begin{minipage}[c]{2.5in}{
\includegraphics[height=2.5in]{#4}
\vspace*{-3mm}\caption[]{#5}\label{#6} \
}\end{minipage}
\end{center}
\vspace*{-0.4in}\
\end{figure*}
}
}

\def\dyfigure[#1,#2,#3,#4,#5,#6]{
{
\begin{figure*}
\vspace*{0.2in}\
\begin{center}
\begin{minipage}[c]{3in}{
\includegraphics[height=3in]{#1} 
\vspace*{-3mm}\caption[]{#2} \label{#3} \
}\end{minipage}\hspace*{0.5in}\
\begin{minipage}[c]{3in}{
\includegraphics[height=3in]{#4} 
\vspace*{-3mm}\caption[]{#5}\label{#6} \
}\end{minipage}
\end{center}
\vspace*{-0.4in}\
\end{figure*}
}
}

\def\dyoldfigure[#1,#2,#3,#4,#5,#6]{
{
\begin{figure*}
\vspace*{0.2in}\
\begin{center}
\begin{minipage}[c]{3in}{
\epsfysize=2.0in\
\hspace{0.5in}\
\epsfbox{#1}
\vspace*{-3mm}\caption[]{#2} \label{#3} \
}\end{minipage}\hspace*{0.25in}\
\begin{minipage}[c]{3in}{
\epsfysize=2.0in\
\hspace{0.5in}\
\epsfbox{#4}
\vspace*{-3mm}\caption[]{#5}\label{#6} \
}\end{minipage}
\end{center}
\vspace*{-0.4in}\
\end{figure*}
}
}

\def\cfiguredouble[#1,#2,#3,#4]{
\begin{figure}
\vspace*{0.2in}\
\begin{center}
\begin{minipage}[c]{1.5in}{
\epsfxsize=1.5in\
\epsfbox{#1}
}\end{minipage}\hspace*{0.1in}\
\begin{minipage}[c]{1.5in}{
\epsfxsize=1.5in\
\vspace{0.1in}\epsfbox{#2}
}\end{minipage}\vspace*{-0.10in} \caption[]{#3}\label{#4}
\end{center}
\vspace*{-0.4in}\
\end{figure}
}

\def\wpfigure[#1,#2,#3,#4]{
\begin{figure*}
\vspace*{4mm}
\begin{center}

\includegraphics[width=#4]{#1} 

\vspace*{-3mm}\caption[]{#2
} \label{#3}

\vspace*{-5mm}
\end{center}
\end{figure*}}

\def\wprfigure[#1,#2,#3,#4,#5]{
\begin{figure*}
\vspace*{4mm}
\begin{center}

\includegraphics[width=#4, angle=#5]{#1} 

\vspace*{-3mm}\caption[]{#2
} \label{#3}

\vspace*{-5mm}
\end{center}
\end{figure*}}

\def\DoubleFigureWSlide[#1,#2,#3,#4,#5,#6,#7,#8,#9]{
\begin{figure*}
\vspace*{#9}
\begin{center}
\begin{minipage}{#4}
\includegraphics[width=#4]{#1}
\vspace*{-3mm}\caption{#2
}\label{#3}
\end{minipage}
\hspace{2em}
\begin{minipage}{#8}
\includegraphics[width=#8]{#5}
\vspace*{-3mm}\caption{#6
}\label{#7}
\end{minipage}
\vspace*{-5mm}
\end{center}
\end{figure*}
}

\def\DoubleFigureW[#1,#2,#3,#4,#5,#6,#7,#8]{
\begin{figure*}
\vspace*{0in}
\begin{center}
\begin{minipage}{#4}
\includegraphics[width=#4]{#1}
\vspace*{-3mm}\caption{#2
}\label{#3}
\end{minipage}
\hspace{2em}
\begin{minipage}{#8}
\includegraphics[width=#8]{#5}
\vspace*{-3mm}\caption{#6
}\label{#7}
\end{minipage}
\vspace*{-5mm}
\end{center}
\end{figure*}
}

\def\DoubleFigureWHack[#1,#2,#3,#4,#5,#6,#7,#8]{
\begin{figure*}
\vspace*{0in}
\begin{center}
\begin{minipage}{3in}
\includegraphics[width=#4]{#1}
\vspace*{-3mm}\caption{#2
}\label{#3}
\end{minipage}
\hspace{2em}
\begin{minipage}{3in}
\includegraphics[width=#8]{#5}
\vspace*{-3mm}\caption{#6
}\label{#7}
\end{minipage}
\vspace*{-5mm}
\end{center}
\end{figure*}
}

\def\ddcfigure[#1,#2,#3,#4]{
\begin{figure*}
\vspace*{0.2in}\
\begin{center}
\begin{minipage}[c]{3in}{
\includegraphics[height=3in]{#1} 
}\end{minipage}\hspace*{0.5in}\
\begin{minipage}[c]{3in}{
\includegraphics[height=3in]{#2} 
}\end{minipage}\vspace*{-0.10in} \caption[]{#3}\label{#4}
\end{center}
\vspace*{-0.4in}\
\end{figure*}
}

\def\ddcfigureSlide[#1,#2,#3,#4,#5]{
\begin{figure*}
\vspace*{#5}\
\begin{center}
\begin{minipage}[c]{3in}{
\includegraphics[height=3in]{#1} 
}\end{minipage}\hspace*{0.5in}\
\begin{minipage}[c]{3in}{
\includegraphics[height=3in]{#2} 
}\end{minipage}\vspace*{-0.10in} \caption[]{#3}\label{#4}
\end{center}
\vspace*{-0.4in}\
\end{figure*}
}

\def\cxfigure[#1,#2,#3]{
\begin{figure}
\vspace*{4mm}
\begin{center}
 
\epsfxsize=2.5in\
\epsfbox{#1}\
 
\vspace*{-0.10in}\caption[]{#2
} \label{#3}
 
\vspace*{-5mm}
\end{center}
\vspace*{-2mm}
\end{figure}}

\usepackage{xcolor}
\definecolor{commentgreen}{RGB}{2,112,10}
\definecolor{eminence}{RGB}{108,48,130}
\definecolor{weborange}{RGB}{255,165,0}
\definecolor{frenchplum}{RGB}{129,20,83}
\usepackage{listings}
\usepackage{amssymb,amsthm}

\icmltitlerunning{\trick: Efficient Intercept Support for Augmented LLM Inference}

\begin{document}

\setlength{\abovedisplayskip}{1pt}
\setlength{\belowdisplayskip}{1pt}
\setlength{\abovedisplayshortskip}{1pt}
\setlength{\belowdisplayshortskip}{1pt}




\twocolumn[
\icmltitle{\trick: Efficient Intercept Support for \\ Augmented Large Language Model Inference}

\icmlsetsymbol{equal}{*}

\begin{icmlauthorlist}
\icmlauthor{Reyna Abhyankar}{equal,sch}
\icmlauthor{Zijian He}{equal,sch}
\icmlauthor{Vikranth Srivatsa}{sch}
\icmlauthor{Hao Zhang}{sch}
\icmlauthor{Yiying Zhang}{sch}
\end{icmlauthorlist} 

     

\icmlaffiliation{sch}{University of California, San Diego, La Jolla, United States}

\icmlcorrespondingauthor{Reyna Abhyankar}{vabhyank@ucsd.edu}
\icmlcorrespondingauthor{Zijian He}{zih015@ucsd.edu}
\icmlcorrespondingauthor{Yiying Zhang}{yiying@ucsd.edu}

\icmlkeywords{Machine Learning Systems, LLM Inference, Augmented LLM}

\vskip 0.3in

]

\printAffiliationsAndNotice{\icmlEqualContribution}





\thispagestyle{empty}


\begin{abstract}
Large language models are increasingly integrated with external environments, tools, and agents like ChatGPT plugins to extend their capability beyond language-centric tasks. However, today's LLM inference systems are designed for standalone LLMs. They treat each external interaction as the end of LLM generation and form a new request when the interaction finishes, causing unnecessary recomputation of already computed contexts, which accounts for 37-40\% of total model forwarding time.
This paper presents \trick, the first LLM inference framework targeting augmented LLMs and supporting the efficient interception of LLM generation. \trick\ minimizes the GPU resource waste caused by LLM interceptions and dedicates saved memory for serving more requests.
\trick\ improves the overall serving throughput by 1.6\x{}-2\x{} and completes 2\x{} more requests per second compared to the state-of-the-art LLM inference systems.


\end{abstract}
\section{Introduction}
\label{sec:intro}

Large language models (LLMs)~\cite{openai2023gpt4,touvron2023llama} have shown immense potential in various applications, such as natural language understanding and content generation. 
Nonetheless, LLMs alone can only generate text. To extend LLMs' capabilities to handle more diverse and open-ended tasks, a recent trend is to augment LLMs with external tools ~\cite{mialon2023augmented, Wang2023ASO, qin2023tool} such as arithmetic calculation~\cite{hao2023toolkengpt} and ChatGPT plugins~\cite{chatgpt-plugin}, non-LLM models like image generation~\cite{betker2024improving}, and interaction with humans and other environments in real-time, such as virtual environments~\cite{shridhar2020alfworld}. 
Notably, Ng poses that ``AI agent workflows will drive massive AI progress in 2024''~\cite{ng-blog}.
More generally, we can view the use of tools and the interaction with humans or other environments all as LLMs being augmented and {\em intercepted} by external entities.

Augmentations have unique properties that bring new problems to LLM inference systems. 
First, they {\em intercept} regular LLM decoding, \ie, when an external tool is called or when waiting for human/environment responses, the normal decoding phase is paused and can only resume when the interception finishes.
Second, an interception's time varies drastically across augmentation types and requests. For example, a calculator API finishes in less than 1\,ms, yet it may take humans more than 1 minute to digest and type a subsequent chat prompt. This complicates request scheduling.
Third, unlike regular requests, the context (\ie, KV caches) cannot be used for a paused request during interceptions but will be needed upon the end of interceptions. It is unclear how to handle {\em temporarily unused context}. Adding to the complexity is the high variation in context length and interception time across requests. 



Unfortunately, existing LLM inference systems~\cite{vLLM,Orca,li2023alpaserve,aminabadi2022deepspeed,nvidia2023fastertransformer,chunk_prefill_sarathi} fall short in serving augmented LLMs with interceptions.
They typically interpret an interception as a termination signal -- they conclude the ongoing request when the interception happens, discard its context, relegate the task to the augmenting entity, and subsequently re-initiate a new request upon receiving the response from the entity.
Consequently, when a request frequently triggers interceptions, for each interception, the initial prompt and all tokens generated before the interception are formed as a new request submitted to the inference system, causing substantial recomputation of keys and values (known as the KV cache) for all previously processed tokens. This recomputation wastes GPU resources, which could instead serve new requests. 

In contrast to the aforementioned interception handling approach (which we refer to as \discard), 
an alternative approach is to preserve a request's context during the interception and resume the request when the interception finishes. This approach, which we call {\em \preserve}, avoids recomputation but occupies GPU memory during the entire interception. This GPU memory could accommodate other requests, which increases throughput~\cite{vLLM}. 

To avoid recomputation and memory preservation, another possible approach to handle interceptions is to swap contexts to CPU memory when an interception happens (we call it {\em \swap}). Although \swap\ avoids recomputation and GPU memory wastage, with limited GPU-CPU link bandwidth, foreground tasks (normal forwarding) could be bottlenecked by waiting for swapping to finish. The GPU memory could be used for processing more requests. 

\textit{How should an LLM inference system efficiently support interceptions?} To answer this question, we first study the computational patterns of six typical augmentations: arithmetic, question-and-answer, virtual environment, chatbot, image generation, and text-to-speech transformation.
We then evaluate how well the state-of-the-art LLM inference system, vLLM~\cite{vLLM} (which performs \discard), and the two possible extensions (\preserve\ and \swap) perform on these augmentations. 


Based on our findings, we propose {\em \trick}, the {\em first} LLM inference framework targeting augmented LLMs with interceptions. 
The core idea of \trick, which we call {\em min-waste interception}, is to minimize GPU memory waste caused by interceptions so that the same amount of GPU resources can be used to serve more requests.
To this end, \trick\ incorporates three contributions.
First, we deduce three waste-calculation equations to quantify the GPU memory waste of \discard, \preserve, and \swap. 

Second, we improve individual interception techniques to reduce or eliminate their memory waste.
For \swap, we propose swap pipelining, which overlaps swap and foreground computation in a model-layer-by-layer manner. Additionally, we split the swapping of a sequence into multiple model-forwarding iterations 
so that each iteration's swapping needs stays within what the GPU-CPU link can sustain (\ie, a swap budget). As a result, \trick\ completely eliminates GPU memory waste of \swap.
For \discard, we observe that certain GPU processing capacity is unused by running requests at each iteration because of decoding's imbalanced memory/compute requirements. 
We utilize this unused capacity for recomputation.
We split requests' contexts into chunks, each small enough to fit in the unused capacity and recomputed in one iteration.

Finally, \trick\ dynamically chooses the interception and resumption strategies within and across requests to minimize overall GPU memory waste while ensuring fairness.
For interception scheduling, we sort intercepted requests by their potential memory waste 
and assign the swap budget to requests with the highest potential waste. 
We discard or preserve the remaining requests' context based on the waste comparison.
For resumption scheduling, we assign the swap budget to swapped-out requests with the earliest original arrival times.
We also schedule discarded requests, preserved requests, and non-intercepted waiting requests according to FCFS (first-come first-served), adding requests until the GPU's processing capacity is reached.
We implement \trick on top of vLLM~\cite{vLLM}, a state-of-the-art LLM inference system. We evaluate \trick, vLLM (\discard), \preserve, and \swap\ on A100 GPUs using three LLMs (GPT-J-6B~\cite{gpt-j-6b}, Vicuña-13B~\cite{vicuna_share_gpt}, and Llama3-70B~\cite{llama3}) and the six interception types we study.
Overall, \sys\ sustains 1.6\x{}-2\x{} higher serving load than vLLM while maintaining similar latency per token generation. \sys\ also achieves over 2\x{} more completed requests per second.

\trick\ is available at \url{https://github.com/WukLab/InferCept}.
\section{Augmented LLMs}
\label{sec:study}

This section first discusses popular augmented LLM frameworks and then presents properties of typical augments from our empirical study.

\subsection{Augemented LLM Frameworks}


To extend LLMs' ability to undertake more types of tasks, different approaches have been proposed to {\em augment} an LLM with various external entities~\cite{mialon2023augmented}. Below, we describe typical augmentations in three categories.

The first type of augmentation involves non-LLM tools called during the decoding phase of an LLM, usually aimed at extending the types of tasks the LLM can handle. These tools range from simple tools like calculator~\cite{wolfram-chatgpt} and calendar
to real-world interactions like information retrieval~\cite{baeza1999modern} and restaurant reservation~\cite{opentable-chatgpt}. A tool augmentation can also be another machine-learning model such as translation~\cite{nllbteam2022language}, and question-answering (QA)~\cite{izacard2022atlas}.
Previous research~\cite{parisi2022talm,hao2023toolkengpt,schick2023toolformer} has shown the effectiveness of fine-tuning an LLM to generate appropriate tool calls automatically during the decoding phase.
Instead of fine-tuning, another approach is instructing LLMs to select tools or models for a user task~\cite{shen2023hugginggpt, lu2023chameleon}.

The second type augments a single triggering of an LLM by making a {\em sequence} of calls to the same LLM. For example, to continuously interact with a human, a chatbot can take rounds of back-and-forth chats by calling an LLM at each step and maintaining a chat history~\cite{greyling-chatgpt-api}. 
Another popular use case is decomposing a complex task into multiple steps. For example, Chain-of-Thought~\cite{wei2022chain,zhang2023automatic} breaks one request into a chain of ``thought'' steps, each being a new prompt to an LLM that continues the history of      thoughts. Similarly, ReAct~\cite{yao2022react} uses a chain of reasoning and action steps to accomplish complex tasks. 

The third type allows for more complex compositions of LLMs, other models, and tools. 
Frameworks such as LangChain~\cite{Chase_LangChain_2022}, DSpy~\cite{khattab2024dspy}, Gorilla~\cite{patil2023gorilla}, SGLang~\cite{zheng2023efficiently}, and AgentGraph~\cite{chen2019agentgraph} provide programming models for users to write their own flows of augmented LLMs.
Other efforts enable LLMs to 
generate compositions of augmented LLMs
~\cite{surís2023vipergpt, qian-etal-2023-creator}.





\subsection{Augmenting Entities and Their Properties}



To understand augmented LLM interceptions and design inference systems for them, we first examine a set of representative augmentations, including arithmetic (math), question-and-answer (QA), virtual environment (VE), chatbot, image generation, and text-to-speech (TTS), to answer three key questions: 1) how long does an interception take? 2) how many interceptions occur for a request? and 3) what is the context length when an interception happens? 
Table~\ref{tbl-api-study} summarizes our results with averages and variations. 
We leave other metrics, such as interception returned token length, detailed CDF distributions, and our evaluation methodology, in the Appendix.
%

\begin{table}[t]
\begin{center}
\footnotesize
\resizebox{\columnwidth}{!}{%
\begin{tabular}{ |c|c|c|c| } 
 \hline
{\bf Type} & {\bf Int Time (sec)} & {\bf Num Interceptions} & {\bf Context Len}\\ 
 \hline

{\bf Math} & (9e-5, 6e-5) & (3.75, 1.3) & (1422, 738)\\
{\bf QA} & (0.69, 0.17) & (2.52, 1.73) & (1846, 428)\\
{\bf VE} & (0.09, 0.014) & (28.18, 15.2) & (2185, 115)\\

{\bf Chatbot} & (28.6,15.6)* & (4.45,1.96) & (753, 703) \\
{\bf Image} & (20.03, 7.8)$\dagger$ & (6.91,3.93)* & (1247,792)\\
{\bf TTS} & (17.24,7.6)$\dagger$ & (6.91,3.93)* & (1251,792)\\
 \hline
\end{tabular}
}
\mycaption{tbl-api-study}{Interception Properties}
{Each cell shows (mean, variance) of the augment. 
* estimated behavior. $\dagger$ partially estimated.
}
\end{center}
\end{table}

\boldunderpara{Arithmetic (Math).}~~~ 
To solve complex math problems, an LLM can be augmented to deconstruct the problem and call a calculator tool step-by-step. We evaluate this use case with the GSM8K-XL~\cite{hao2023toolkengpt} dataset, which contains 8.5K high-quality grade-school math problems.
As expected, the calculator's execution times are short, an average of 0.2 ms. It has a fairly large context length when calculators are called since we must prompt the LLM with demonstrations of solving problems step-by-step. 

\boldunderpara{Knowledge-based question and answer (QA).}~~~ 
To allow pre-trained LLMs to access wider sets of knowledge, one can augment the LLM to call knowledge-based QA tools that retrieve information from a rich dataset.
We use the Multihop QA Wikipedia~\cite{yang2018hotpotqa} dataset to evaluate this use case. 
While having longer context lengths for questions, these QA tools provide relatively quick yet variable execution times due to the non-deterministic network communication latency with Wikipedia.


\boldunderpara{Virtual Environment (VE).}~~~ 
An LLM can be augmented to interact with virtual environments and accomplish complex tasks by performing gradual steps~\cite{Gu_2022, huang2022language, hu2023look}. 
To evaluate VE, we use the ALFWorld dataset~\cite{shridhar2020alfworld}, an interactive environment that aligns text descriptions and commands with a physically embodied robotic simulation.
The VE interaction has a short and stable interception time due to a locally executed and embodied text-based environment. 
The context length is long due to the instructions needed to prompt the LLM to understand VE action sequences.

\boldunderpara{Chatbot.}~~~ 
A popular task for LLMs is a chatbot, which interacts with humans with a sequence of chats. When a human receives a chat response from the LLM, the LLM generation sequence is essentially intercepted. The LLM generation is resumed when the human sends the subsequent chat message. During this process, the chat history must be kept as the context for the entire chat sequence. 
We use the ShareGPT dataset~\cite{vicuna_share_gpt}, which contains crowd-sourced real-user ChatGPT conversations to evaluate chatbots. 
We estimate human response time (\ie, the interception time) as the summation of human scanning time~\cite{brysbaert_2019} of the previous chat response and typing time~\cite{ma2015haptic} of the next prompt. 
The context 
lengths vary greatly due to the variety of human prompts. 
Its interception time also has a high variance.

\boldunderpara{Image generation.}~~~ 
An LLM can be augmented to generate an image or a sequence of images when a human gradually refines features through multiple prompts (\eg, adding details to a face depiction). 
This involves two interceptions: one for triggering an image-generation model and another for receiving user subsequent prompts. 
To understand this use case, we use ChatGPT to create a dataset by generating a series of image-generation prompts, each triggering a call to the Stable Diffusion model~\cite{stable_diffusion}. 
This workload's interception time has high variation due to variations in executing the diffusion model and the variation in human response time. 
The average context length is shorter than the chatbot interception due to smaller input prompts to the diffusion model. 

\boldunderpara{Text-To-Speech (TTS).}~~~ 
As demonstrated by the ChatGPT Whisper API support~\cite{openai-chatgpt-whisper-apis}, TTS can be integrated with an LLM to communicate with a human. 
Similar to our image-generation dataset, we use ChatGPT to generate a series of prompts, each triggering a call to the Bark TTS model~\cite{tts_sunoai2023bark}. Similar to image generation, we estimate 
the human response time but measure actual TTS execution time. 
TTS's behavior also has high variation in interception time due to variations in model execution time and user response time.



\boldunderpara{Summary and insights.}~~~
Overall, we find LLM interception time is highly dependent on augmentation types and exhibits a clear difference between short-running (Math, QA, VE) and long-running (Chatbot, Image, TTS) ones. The short-running augmentations are fully automated with no human interaction. Most also exhibit small variations (except for QA, which interacts with the network). 
Augmentations that involve human interaction and/or call another large model are long-running with high variations. 
The short/long interception times and their variation imply that no single LLM interception strategy is universally applicable, but an inference system could utilize augmentation type as a hint for interception time.
Context lengths for all the augmentations are large, implying potentially high GPU memory wastage.
Context lengths also see high variation, complicating the handling of them at runtime.


{
\begin{figure*}[th]
\begin{center}
\centerline{\includegraphics[width=\textwidth]{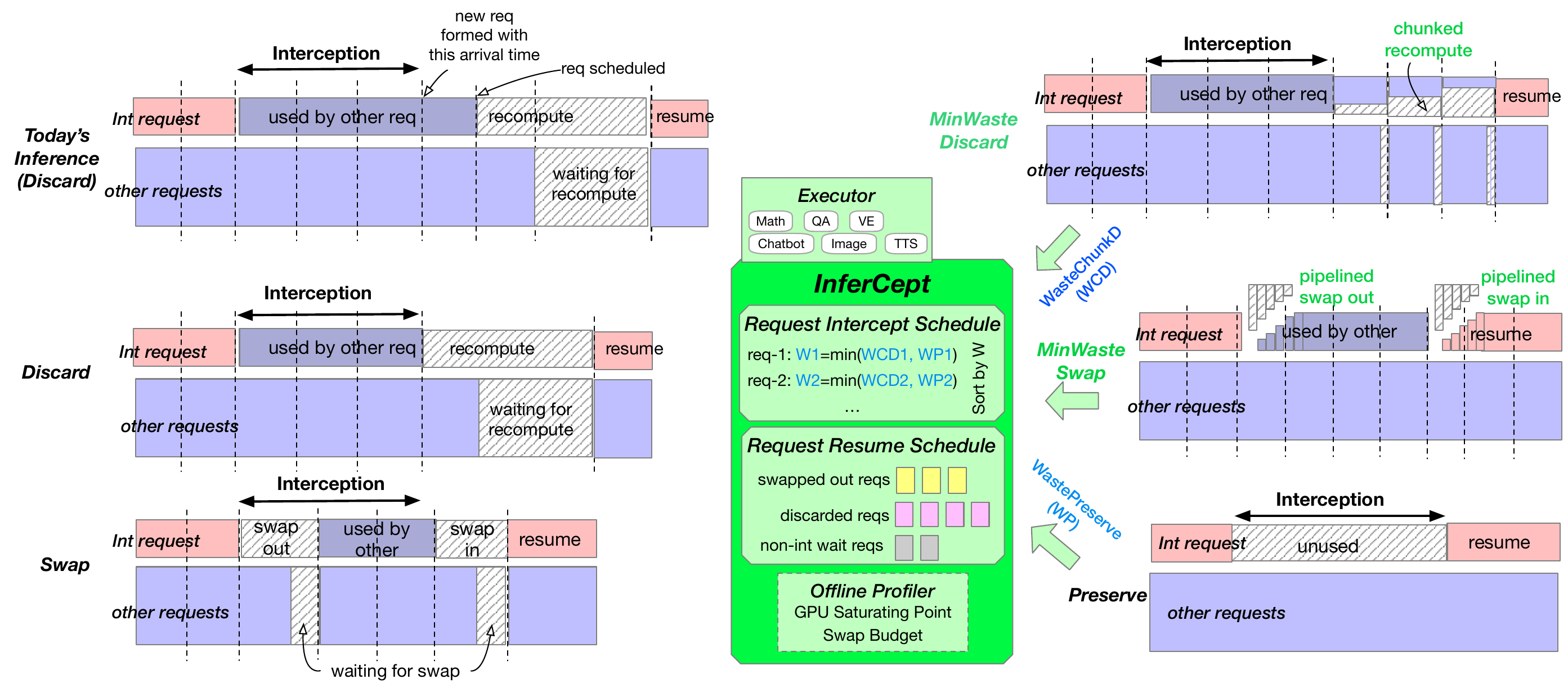}}
\mycaption{fig-existing}{\trick\ and Alternative Approaches.}
{
Vertical: GPU resources occupied or wasted by intercepted requests and other normal ones.
Horizontal: timeline divided by iterations (dotted vertical lines). 
Hatch parts represent memory waste.
\trick\ adaptively combines preserve, our optimized discard, and our optimized swap.
All green parts represent \trick's techniques.
}
\end{center}
\end{figure*}
}

\section{Existing LLM Inference Systems}
\label{sec:inference}

This section discusses related work in LLM inference systems and why simple extensions to them are insufficient for interception handling, as shown in Figure~\ref{fig-existing}.

\subsection{LLM Inference Systems}
\label{sec:inference-sys}

Recent years have seen the rise of inference systems that are designed specifically for LLMs. 
Unlike \trick, none of the existing inference systems are designed for handling interceptions, and all suffer from unoptimized performance, as we will show in Section~\ref{sec:existing-study}.

Some works aim to improve overall inference throughput with better scheduling strategies. For example, 
Orca~\cite{Orca} proposes iteration-level scheduling where, at the end of each model forward pass, Orca's scheduler is invoked to form a new batch for the next forward pass. Such iteration-level scheduling strategies improve GPU utilization, resulting in higher inference throughput. 
\trick\ also adopts iteration-level scheduling but with a unique and new scheduling strategy --- at each iteration, \trick\ makes a decision that minimizes GPU memory wastage for requests intercepted or resumed in that iteration based on their interception properties.
Besides FCFS, which Orca and others use, alternative request scheduling policies such as a multi-level feedback queue in FastServe~\cite{wu2023fast} have been proposed. We leave the comparison of these scheduling policies to future work.

Another optimization target is GPU memory usage. vLLM~\cite{vLLM} proposes the concept of paged attention, where the KV cache is treated as virtual memory that can map to non-contiguous physical GPU memory. Because of the flexibility in virtual memory, vLLM achieves better utilization of GPU memory, thereby improving overall inference performance. \trick\ also utilizes paged attention to efficiently use GPU physical memory. Unlike vLLM, which is agnostic to interceptions, \trick\ adaptively preserves, discards, or swaps the KV cache of intercepted requests to minimize memory waste. Other memory-efficient techniques, such as prefix sharing in SGLang~\cite{zheng2023efficiently} and Preble~\cite{srivatsa2024preble}, are orthogonal and can be added to \trick.

Yet another set of LLM inference systems focuses on the imbalanced computation needs between the prefill and decoding stages. Sarathi~\cite{chunk_prefill_sarathi} is a system that proposes the technique of chunked prefill, which splits prompt tokens into chunks, each merged with other decoding requests to form one iteration's batch. 
\trick's chunking splits sequences for recomputation and swapping to fully utilize GPU and GPU-CPU-link resources while minimizing memory waste.
Other proposals on prefill-decoding optimization like DistServe~\cite{zhong2024distserve} are orthogonal and can potentially be added to \trick.

\subsection{Limitations of Inference Systems and Extensions}
\label{sec:existing-study}

We now qualitatively and quantitatively analyze the limitations of existing LLM inference systems and naive extensions to them to support interceptions. We quantify the impact of these approaches on serving speed with our proposed {\em GPU memory waste} calculations.

\boldunderpara{Discard-based approaches.}~~~
Today's LLM inference systems 
are not designed for augmented LLMs.
To serve LLMs with augmentations, such a system would need to treat the interception as the end of the request and re-initiate a request when the interception finishes. As today's inference systems use the FCFS scheduling policy, they schedule these re-initiated requests at the end of the waiting queue and starve resumed requests. 
An easy fix to this scheduling problem is 
using an intercepted request's original arrival time when re-inserting it into the waiting queue (we call this scheme \improvediscard). 

\discard\ and \improvediscard\ both require the recomputation of all context tokens and incur GPU memory waste from two sources.
First, recomputation consumes memory that is not used to produce any new tokens.
Assuming the context of request $i$ when interception $j$ occurs has $C_i^j$ tokens, each token's KV cache occupies $M$ memory, and recomputation takes $T_{fwd}(C_i^j)$ time, where $T_{fwd}$ is a mapping from the number of scheduled tokens in a batch to the execution time of that iteration. The memory wasted for recomputation is $T_{fwd}(C_i^j) \times C_i^j \times M $.
Second, recomputing the context increases iteration time, as an iteration now needs to finish all the recomputations alongside model forwarding of other running requests.
The memory occupied by the other running requests is wasted during the additional iteration time, $T_{fwd}(C_i^j)$. Thus, the memory waste for this reason is $T_{fwd}(C_i^j) \times C_{other} \times M$, where $C_{other}$ is the sum of the contexts of other requests. The total memory waste for \discard\ and \improvediscard\ is: 

\begin{small}
    \begin{equation}
\label{eq-discard-waste}
\begin{split}
    \text{WasteDiscard}_i^j & = T_{fwd}(C_i^j) \times C_i^j \times M \\
    & + T_{fwd}(C_i^j) \times C_{other} \times M \\
\end{split}
\end{equation}
\end{small}

In practice, our evaluation shows that \discard\ incurs a GPU resource wastage (GB*min) of 27\%, and 37-40\% of the total model forwarding time is spent on recomputation with a mixed workload containing all six augmentations in \S\ref{sec:study}. 

\boldunderpara{Preserve-based approach.}~~~
Instead of discarding, one could preserve the context of a request when interception happens. This \preserve\ strategy avoids recomputation cost and can immediately resume a request when the interception finishes. However, the preserved context wastes GPU memory while the request remains paused. 
The preserve waste for request $i$ when interception $j$ occurs is the duration of that interception, $T_{INT}^j$, multiplied by the amount of GPU memory held by the request's context. 

\begin{small}
\begin{equation}
\label{eq-preserve-waste}
\text{WastePreserve}_i^j = T_{INT}^j \times C_i \times M
\end{equation}
\end{small}

Our measured GPU memory waste with real interceptions is surprisingly high: nearly half of the GPU memory is occupied by interrupted requests more than 60\% of the time.

\boldunderpara{Swap-based approach.}~~~
To avoid recomputation overhead in \discard\ and memory waste in \preserve, one potential technique is to swap all context data from GPU to CPU memory when an interception happens and swap it back to GPU when the interception ends. 
The straightforward implementation of \swap\ performs swapping in a synchronous manner by launching CUDA kernels for moving data out/in and letting computation utilize the relevant memory space. Swapping can take an extremely long time with huge contexts because of the volume of data to be moved and the kernels to be launched. To elaborate on the latter, multiple kernels need to be launched, each for a discontinuous physical memory region. With PagedAttention~\cite{vLLM}, the context of an intercepted request can scatter across many physical memory regions, causing high kernel launch overhead. 

Like \discard, \swap\ incurs memory waste from two sources.
First, the memory space being swapped is wasted during swapping, accounting for $T_{swap}(C_i^j) \times C_i^j \times M$, where $T_{swap}$ is a mapping from the number of tokens to swap to the corresponding swapping latency.
Second, \swap\ could increase the iteration time, causing all running requests to wait for swapping to finish and do nothing. This waste is $T_{swap}(C_i^j) \times C_{other} \times M$. The total waste doubles to account for swapping in and out:

\begin{small}
\begin{equation}
\label{eq-swap-waste}
\begin{split}
    \text{WasteSwap}_i^j & = 2 \times T_{swap}(C_i^j) \times C_{batch} \times M
\end{split}
\end{equation}
\end{small}

Our evaluation shows that \swap\ wastes 26\% GPU resources, and over 25\% of the total workload time is spent on waiting for swapping for the mixed workload.
\section{\trick}
\label{sec:design}

This section presents \trick, first with our improved swap and recomputation mechanisms, then with our intercepting and resuming request scheduling, and ending with interception duration estimation and implementation details. Figure~\ref{fig-existing} shows \trick's overall architecture.

\subsection{Swap Pipelining and Chunking}
\label{sec:improve-swap}

We discuss how \trick\ avoids swap memory waste.


\boldunderpara{Swap pipelining and overlapping.}~~~
\swap\ performs swapping in a serial manner and blocks foreground computation for the entire swapping time. To mitigate this problem, we propose to perform swapping in a pipelined manner and in the background.
Specifically, \trick\ views each model layer's swapping as one pipeline stage and pipelines kernel launching, data movement, and foreground model forwarding. For example, when we launch the swap kernel for layer $i+2$, we move the context for layer $i+1$, and layer $i$'s context has been freed and used for normal forwarding. 
%


\boldunderpara{Swap chunking.}~~~
To further reduce memory waste, 
we propose to chunk swap-out and swap-in across multiple iterations so that in each iteration, the swap latency can be hidden by overlapping it with model forwarding. We obtain $T_{fwd}$ through offline profiling and calculate $T_{swap}$ based on the swapping bandwidth and the per-token memory requirements $M$.
At iteration $i$ with batch size $B_i$, we set $T_{swap}(N_i) = T_{fwd}(B_i)$ and calculate $N_i$. $N_i$, what we call the \textit{swap limit}, indicates the number of tokens that can be swapped for free (\ie, hidden behind model forwarding). 



\boldunderpara{Determining swap-in and swap-out budget.}~~~
At every iteration, there can be tokens waiting to be swapped out and swapped in.
\trick\ determines how much bandwidth to give to swap-out and to swap-in at each iteration
by maximizing inference throughput (\ie, the number of tokens that can be added to an iteration's processing) while guaranteeing the following criteria:
1) the total amount of swap-in and swap-out tokens should be at most $N_i$;
2) the amount of swapped-out memory should not exceed the amount of free CPU memory plus the swapped-in memory;
and 3) the amount of swapped-in memory and memory for newly scheduled tokens (\ie, our maximizing target) should not exceed the amount of swapped-out memory and free GPU memory. We use the obtained swap-in and swap-out budget when deciding the request interception schedule in \S\ref{sec:inter-request}.



\subsection{Recomputation Chunking}
\label{sec:improve-discard}

Unlike \swap, which consumes GPU-CPU-link bandwidth and can be hidden from foreground GPU tasks, the recomputation of discarded tokens requires GPU resources that cannot be avoided.
However, we observe that decoding and recomputation have complementary resource requirements, with decoding requiring less GPU core resource (for one query token) than recomputation (queries for the entire context length) for the same amount of memory. Thus, a batch of decoding requests usually cannot fill all GPU cores before running out of GPU memory. Mixing decoding and recomputation requests can increase GPU core utilization while staying within its memory boundary. The main challenge is how much recomputation to mix with decoding.

\discard\ and \improvediscard\ impose the recomputation of an entire request in one iteration. Recomputation of a long context would add more computation burden than what a GPU can handle, causing significantly increased iteration time. As a result, long $T_{fwd}(C_i^j)$ causes huge $WasteDiscard$ in Equation~\ref{eq-discard-waste}.

To mitigate this issue, we propose an adaptive technique that separates the recomputation of a context sequence into chunks, each added to one iteration without going beyond what the GPU can sustain.
We observe that the number of query tokens that all GPU cores can process in parallel is limited and fixed for a given model architecture. We call this number the GPU {\em saturation point}, $S$, expressed as query token count. Processing more query tokens beyond $S$ increases iteration time without improving serving throughput.
\trick\ obtains $S$ from offline profiling and sets the chunk size as $S$ minus the running group size. 

The memory waste of our chunked recomputation mechanism can be calculated in two parts, similar to Equation~\ref{eq-discard-waste}.
For the recomputation itself, we add one chunk of memory waste per iteration that lasts until all chunks have been recomputed, which essentially cuts the waste of \discard's all-recomputation-at-once scheme by half, \ie, $T_{fwd}(C_i^j) \times C_i^j \times M / 2$, as illustrated in Figure~\ref{fig-existing}.
The second part of waste comes from other requests' occupied memory during recomputation-added iteration time. This waste amounts to $n \times T_{fwd}(\frac{C_i^j}{n}) \times C_{other} \times M  $, where $n$ is the number of iterations to finish recomputing the request and $T_{fwd}(\frac{C_i^j}{n})$ is the per-chunk added iteration time.
Adding these two parts, we have the total recomputation memory waste of chunked recomputation as 

\begin{footnotesize}
\begin{equation}
\label{eq-chunkeddiscard-waste}
\begin{split}
    \text{WasteChunkD}_i^j 
             & = \frac{T_{fwd}(C_i^j) \times C_i^j \times M}{2} \\
             & + n \times T_{fwd}(\frac{C_i^j}{n}) \times C_{other} \times M 
\end{split}
\end{equation}
\end{footnotesize}

Compared to Equation~\ref{eq-discard-waste}, the left term (recomputation itself) cuts the corresponding term of \discard\ by half, and the right term (other requests) is no larger than the right term of \discard\ because $n \times T_{fwd}(\frac{C_i^j}{n}) \leq T_{fwd}(C_i^j)$. 
Meanwhile, \trick\ increases GPU core utilization and further improves overall serving throughput.

\subsection{Inter-Request Action Decision}
\label{sec:inter-request}

\boldunderpara{Scheduling intercepted requests.}
We now discuss how \trick\ schedules intercepted requests using a hybrid of preserve, our improved swap, and our improved discard when considering multiple requests together.
First, for each request $i$ encountering interception $j$, we calculate its memory waste as the minimum of WastePreserve (Equation~\ref{eq-preserve-waste}) and WasteChunkDiscard (Equation~\ref{eq-chunkeddiscard-waste}):

\begin{small}
\begin{equation}
\label{eq-waste}
 \text{Waste}_i^j = \textbf{min} (\text{WastePreserve}_i^j,\text{WasteChunkD}_i^j)   
\end{equation}    
\end{small}

Next, we sort all intercepted requests in descending order based on their memory waste. 
We swap out context from these requests according to this order 
until we run out of the swap-out budget.
For the remaining paused requests, we preserve or discard their remaining context based on its decision from Equation~\ref{eq-waste}.



\boldunderpara{Scheduling resumed and other waiting requests.}
At the beginning of each iteration, \sys\ determines what requests to insert into the batch for the iteration.
To facilitate this decision, \sys\ maintains three queues: a running queue containing currently running requests in the system, a swap queue containing requests that have been resumed but are previously swapped out during the interception, and a waiting queue. The waiting queue contains discarded requests that have been resumed, new requests received by \sys\ that have never been served, and previously running requests that have been evicted due to the lack of GPU resources. The latter two types are the same as today's inference systems.
Each queue is sorted by its requests' original arrival times.

At each iteration, we choose requests from the waiting queue in the FCFS order until the GPU saturation point is reached; FCFS ensures fairness and avoids starvation.
Requests chosen can be any of the three aforementioned waiting types. If it is a discarded request, the iteration will perform the recomputation of the scheduled number of tokens. If this request has only been partially recomputed, it remains in the wait queue with the remaining tokens.
At each iteration, we also choose requests from the swap queue in the FCFS order until the swap-in budget is reached. We maintain and schedule a separate swap queue because the swap-in budget is additional to GPU resources and should always be utilized by resumed requests as much as the budget allows.
\subsection{Interception Duration Estimation}
\label{sec:unknown-api}

The calculation of memory waste for preserving context (Equation~\ref{eq-preserve-waste}) requires interception duration.
For interceptions with highly variable durations or those not profiled offline,
we propose a dynamic estimation method by setting $\hat{T}_{INT} = t_{now} - t_{call}$ where $t_{now}$ is the current time updated each iteration and $t_{call}$ is when the last interception was initiated. 
Effectively, the longer a request is intercepted, the larger the estimated interception time.
Our evaluation shows that \trick\ using this estimation method achieves 93\% of the performance compared with using an oracle providing exact interception durations in mixed workload.


\subsection{Implementation}
\sys\ comprises four key components: a scheduling policy, waste calculation, chunked recomputation and swapping, and diverse augmentation support. We implement \sys\ on top of vLLM~\cite{vLLM} to leverage its PagedAttention technique for regular LLM memory management, Most of our techniques are highly modular and orthogonal to optimizations designed for non-intercepted LLMs. Thus, \sys\ can be potentially integrated into other LLM serving systems like DeepSpeed~\cite{aminabadi2022deepspeed}, Orca~\cite{Orca}, and TensorRT-LLM~\cite{tensorRTllm}.

{
\begin{figure*}[t]
\begin{center}
\centerline{\includegraphics[width=\textwidth]{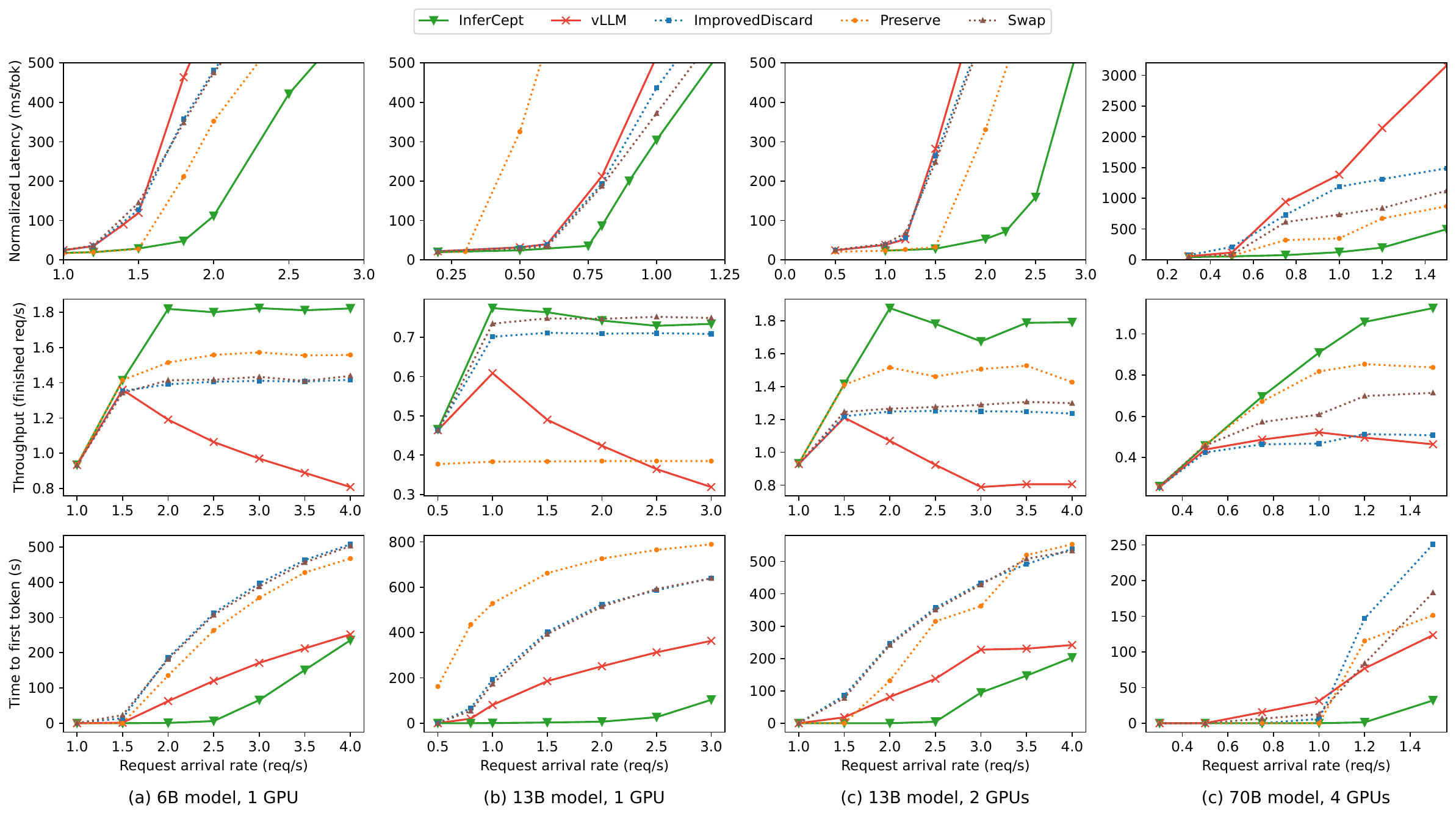}}
\mycaption{fig-e2e-all}{End-to-end Performance on Mixed Workload.}
{
\textbf{First Row: Normalized Latency.} Lower right is better, i.e., sustains higher serving load.
\textbf{Second Row: Throughput.} Expressed as completed requests per second. Higher is better.
\textbf{Third Row: Time-to-first-token (TTFT).} Lower is better, i.e., shorter response time. 
}
\end{center}
\end{figure*}
}

\section{Evaluation Results}
\label{sec:results}

We compare \sys\ to vLLM (with no modifications), \improvediscard, \preserve, and \swap, as discussed in \S\ref{sec:existing-study}.
To evaluate \sys\ and the baselines, we augment the 6B-parameter GPT-J model~\cite{gpt-j-6b}, the 13B Vicuna model~\cite{vicuna_share_gpt}, and the 70B Llama3 model~\cite{llama3}
with workloads presented in \S\ref{sec:study}. We run augmented GPT-J on one NVIDIA A100 GPU. For Vicuna, we use two environments: running on a single A100 GPU and distributed on two A100 GPUs with tensor parallelism. For Llama3, we distribute it on four A100 GPUs with tensor parallelism. 
To mimic real-world serving scenarios that often receive different types of requests, we use a request dataset that merges the six augmentations presented in \S\ref{sec:study} by uniformly sampling requests from them. 

\subsection{End-to-End Performance}
\label{sec:e2e-exps}

We first compare the end-to-end performance of \sys\ and the baselines.
Following recent LLM inference research papers~\cite{vLLM,Orca}, we first report the serving throughput as normalized latency (\ie, the median of every request's end-to-end latency divided by its output length) when varying request load (number of requests arrived per second). We also remove a request's intercepted time from its end-to-end latency because it is the same across all serving systems. Under the same normalized latency, a higher-throughput system should sustain a higher request rate.
Apart from normalized latency, we also report the number of finished requests per second (\ie, throughput) and the time from request arrival to the first generated token (\ie, TTFT). These metrics are important in meeting service-level objectives and system response time~\cite{zhong2024distserve, Hu2024InferenceWI, patel2024splitwise}.
Figure~\ref{fig-e2e-all} shows these results for \sys\ and the baselines on all three models. For Vicuna-13B, we report both a single GPU and two GPU executions.

\boldunderpara{6B model results.}~~~
\sys\ outperforms baselines across all metrics. \sys\ sustains up to 1.6\x{} higher request arrival rate at the same normalized latency compared to vLLM. Under the same request rate, \sys's normalized latency is 1.9\x{}-5.7\x{} lower than vLLM. \sys\ serves 1.7\x{} higher request arrival rate than vLLM with the smallest TTFT. 
\sys\ outperforms all of the baselines mainly because of our min-waste-based schedule. 
Moreover, \sys\ improves the recomputation and swap mechanisms over \improvediscard\ and \swap\ with chunking and pipelining. Compared to them, \sys\ eliminates over 60\% of GPU waste caused by recomputation and 96\% of GPU waste for swapping.

The vanilla vLLM suffers from high recomputation waste and delays in scheduling resumed requests per FCFS policy, resulting in worse throughput and normalized request latency. Its TTFT is better than other baselines and close to \sys\ because it puts intercepted requests to the end of the wait queue, which is a round-robin schedule. This results in the system primarily handling requests before their first interception, essentially allowing TTFT not to be affected by interceptions.
%
%
\improvediscard\ maintains an intercepted request's original arrival time. Thus, both its normalized latency and throughput are better than vanilla vLLM. However, \improvediscard\ still incurs the same recomputation overhead and underperforms \sys\ on all metrics. 
%
\preserve\ is the best among all the baselines regarding normalized latency and throughput, showing that the preserving overhead for our mixed workload with small models is lower than the recomputation overhead for most requests.
Yet, \preserve\ is worse than \sys\ on all metrics because of \sys's dynamic min-waste scheduling and improved swapping and recomputation.
%
The overall results for \swap are similar to \improvediscard. Both baselines avoid GPU memory waste during interceptions. 
While \improvediscard\ causes running requests to wait for the recomputation requests in an iteration, \swap\ causes running requests to wait for swapping out/in requests in an iteration. These overheads have similar effects on the final performance metrics.  


\boldunderpara{13B model results.}~~~
When running the 13B model on a single GPU, \sys\ outperforms all the baselines in normalized latency but with smaller improvements than the 6B model. A larger model's weights occupy more GPU memory, reducing space for the KV cache. 
Moreover, the increased normal forwarding time contributes to most of the normalized latency, limiting the room for improvement.
Still, we serve up to 1.25\x{} higher request rates without a noticeable increase in the normalized latency and also sustain 3.1\x{} higher load with the smallest TTFT.

A more practical scenario for large model inference is a distributed execution environment. We evaluate the 13B model with tensor-parallelism~\cite{shoeybi2020megatronlm} on two GPUs. \sys\ outperforms all baselines by a larger margin than when running on a single GPU. \sys\ serves up to 1.8\x{} higher request arrival rates and a 1.6\x{}-10\x{} improvement in normalized latency compared to vLLM. 
We obtain more benefits in the distributed setting because each GPU uses less memory for storing model weights and allocates more space for the KV cache. This allows more requests to be served concurrently and thus introduces more interceptions. As \sys\ minimizes GPU memory waste from these interceptions, the improvement over baselines becomes more pronounced in this setting.

\boldunderpara{70B model results.}~~~
For the Llama3-70B model, \sys\ has substantial improvement over all baselines. \sys\ can sustain 2\x{} higher request arrival rate and achieve 1.3\x{}-12\x{} lower normalized latency at the same RPS. We also achieve 2.4\x{} higher load than vLLM with the same TTFT.
As model size increases, both vanilla vLLM and \improvediscard experience significant recomputation costs. In contrast, \preserve and \swap perform better. This is attributed to the adoption of grouped query attention (GQA)~\cite{ainslie-etal-2023-gqa}, which compresses the attention KVs. This mechanism reduces inactive memory for \preserve and moderates data movement for \swap. By making optimal preemption decisions tailored to specific requests, \sys\ provides more benefits since the GPU waste is lowered significantly.

\boldunderpara{Single-augment workloads.}~~~
Apart from the mixed workload, we evaluate \sys\ on two single-API workloads: QA and Chatbot (\S\ref{sec:study}). \sys\ outperforms vLLM by up to 2.3\x{} and 1.9\x{} on normalized latency with QA and Chatbot respectively. \sys's improvement with QA is larger because most QA API calls are short and have smaller waste when using preserve than discard.

\subsection{Performance Deep Dive}

To further understand \sys's benefits, we break down its techniques by adding them one at a time to vanilla vLLM. Using the mixed workload, we report the normalized latency and GPU memory waste under the load of 2 requests per second for the 6B model in Figure~\ref{fig-ablation}. Other request loads have similar trends.

{
\begin{figure}[t]
\begin{center}
\centerline{\includegraphics[width=\columnwidth]{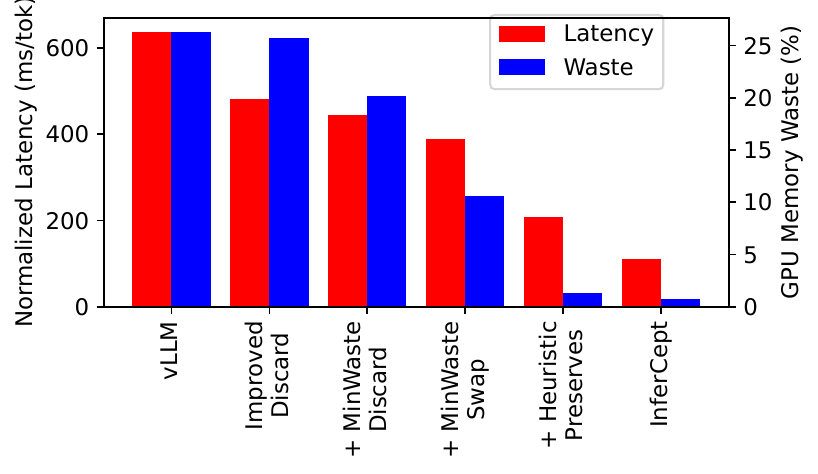}}
\vspace{-0.05in}
\mycaption{fig-ablation}{\sys\ Technique Breakdown.}
{
Each bar group adds one technique over its left bar group, with the leftmost being vanilla vLLM and the rightmost being the full \sys.
}
\end{center}
\end{figure}
}

We first improve \discard\ (vanilla vLLM) by maintaining the request's original arrival time as in \improvediscard, which reduces normalized latency by 24.5\%. We then add \trick's recomputation chunking, resulting in 7.8\% more improvement. Next, we add \trick's budgeted swapping (discarding once the limit is reached). This results in an added improvement of 12.7\%. Then, we add preserve and use a simple heuristic to decide between discard and preserve: discard for interactive (or long-running) interceptions and preserve for automated (short-running) interceptions. This addition has a 46.1\% improvement, mainly because of the added support for \preserve. Finally, adding min-waste-based adaptive schedule (\ie, the whole \trick) results in 
an additional 46.4\% improvement.
Each of \trick's techniques reduces GPU memory waste, with the whole \trick\ only having 0.69\% waste. This demonstrates the effectiveness of our min-waste preemption principle. 

Note that \sys's techniques' benefits differ for different workloads. For example, our chunking and pipelining techniques for recomputation and swapping contribute 54\% of total speedup for the single Chatbot workload. This is because Chatbot has a long interception duration, allowing for more requests to execute and, in turn, triggering more swap and recomputation.
\section{Conclusion}
We introduced \trick, an LLM inference framework optimized for interceptions during inference. By minimizing GPU memory waste as a unified objective, 
\trick delivers 1.6\x{}-2\x{} higher request arrival rates and achieves 1.3\x{}-12\x{} lower normalized latency compared to state-of-the-art inference systems.

\section*{Acknowledgements}
We would like to thank the anonymous reviewers for their tremendous feedback and comments, which have substantially improved the content and presentation of this paper. 
We are also thankful to Longfei Yun, Shravan Konduru, Hao-Ting Tso, Chujia Guo, Yu-shu Chen, Chace Zhu, Shivam Pansuria, and Katherine Guo for their contribution to the initial phase of this work.
This material is based upon work supported by gifts from AWS, Google, and Meta. Any opinions, findings, conclusions, or recommendations expressed in this material are those of the authors and do not necessarily reflect the views of these institutions.

\section*{Impact Statement}

This paper presents work whose goal is to advance the field of Machine Learning. There are many potential societal consequences of our work, none of which we feel must be specifically highlighted here.


\bibliographystyle{icml2024}
\bibliography{references}

\begin{thebibliography}{56}
\providecommand{\natexlab}[1]{#1}
\providecommand{\url}[1]{\texttt{#1}}
\expandafter\ifx\csname urlstyle\endcsname\relax
  \providecommand{\doi}[1]{doi: #1}\else
  \providecommand{\doi}{doi: \begingroup \urlstyle{rm}\Url}\fi

\bibitem[Achiam et~al.(2023)Achiam, Adler, Agarwal, Ahmad, Akkaya, Aleman, Almeida, Altenschmidt, Altman, Anadkat, et~al.]{openai2023gpt4}
Achiam, J., Adler, S., Agarwal, S., Ahmad, L., Akkaya, I., Aleman, F.~L., Almeida, D., Altenschmidt, J., Altman, S., Anadkat, S., et~al.
\newblock Gpt-4 technical report.
\newblock \emph{arXiv preprint arXiv:2303.08774}, 2023.

\bibitem[Agrawal et~al.(2023)Agrawal, Panwar, Mohan, Kwatra, Gulavani, and Ramjee]{chunk_prefill_sarathi}
Agrawal, A., Panwar, A., Mohan, J., Kwatra, N., Gulavani, B.~S., and Ramjee, R.
\newblock Sarathi: Efficient llm inference by piggybacking decodes with chunked prefills.
\newblock \emph{arXiv preprint arXiv:2308.16369}, August 2023.

\bibitem[AI(2023)]{tts_sunoai2023bark}
AI, S.
\newblock Bark: Text-to-speech model.
\newblock \url{https://github.com/suno-ai/bark}, 2023.

\bibitem[Ainslie et~al.(2023)Ainslie, Lee-Thorp, de~Jong, Zemlyanskiy, Lebron, and Sanghai]{ainslie-etal-2023-gqa}
Ainslie, J., Lee-Thorp, J., de~Jong, M., Zemlyanskiy, Y., Lebron, F., and Sanghai, S.
\newblock {GQA}: Training generalized multi-query transformer models from multi-head checkpoints.
\newblock In \emph{Proceedings of the 2023 Conference on Empirical Methods in Natural Language Processing (EMNLP 2023)}, Singapore, December 2023.

\bibitem[Aminabadi et~al.(2022)Aminabadi, Rajbhandari, Awan, Li, Li, Zheng, Ruwase, Smith, Zhang, Rasley, et~al.]{aminabadi2022deepspeed}
Aminabadi, R.~Y., Rajbhandari, S., Awan, A.~A., Li, C., Li, D., Zheng, E., Ruwase, O., Smith, S., Zhang, M., Rasley, J., et~al.
\newblock Deepspeed-inference: enabling efficient inference of transformer models at unprecedented scale.
\newblock In \emph{SC22: International Conference for High Performance Computing, Networking, Storage and Analysis}, Dallas, Texas, November 2022. IEEE.

\bibitem[Baeza-Yates et~al.(1999)Baeza-Yates, Ribeiro-Neto, et~al.]{baeza1999modern}
Baeza-Yates, R., Ribeiro-Neto, B., et~al.
\newblock \emph{Modern Information Retrieval}, volume 463.
\newblock ACM Press, New York, 1999.

\bibitem[Betker et~al.(2024)Betker, Goh, Jing, Brooks, Wang, Li, Ouyang, Zhuang, Lee, Guo, Manassra, Dhariwal, Chu, Jiao, and Ramesh]{betker2024improving}
Betker, J., Goh, G., Jing, L., Brooks, T., Wang, J., Li, L., Ouyang, L., Zhuang, J., Lee, J., Guo, Y., Manassra, W., Dhariwal, P., Chu, C., Jiao, Y., and Ramesh, A.
\newblock Improving image generation with better captions.
\newblock \url{https://cdn.openai.com/papers/dall-e-3.pdf}, 2024.

\bibitem[Brockman et~al.(2023)Brockman, Eleti, Georges, Jang, Kilpatrick, Lim, Miller, and Pokrass]{openai-chatgpt-whisper-apis}
Brockman, G., Eleti, A., Georges, E., Jang, J., Kilpatrick, L., Lim, R., Miller, L., and Pokrass, M.
\newblock Introducing chatgpt and whisper apis.
\newblock \url{https://openai.com/blog/introducing-chatgpt-and-whisper-apis}, March 1 2023.

\bibitem[Brysbaert(2019)]{brysbaert_2019}
Brysbaert, M.
\newblock How many words do we read per minute? a review and meta-analysis of reading rate.
\newblock \emph{Journal of memory and language}, 109:\penalty0 104047, 2019.

\bibitem[Chase(2022)]{Chase_LangChain_2022}
Chase, H.
\newblock {LangChain}.
\newblock \url{https://github.com/langchain-ai/langchain}, October 2022.

\bibitem[Chen et~al.(2019)Chen, Chen, Tan, Long, Gasic, and Yu]{chen2019agentgraph}
Chen, L., Chen, Z., Tan, B., Long, S., Gasic, M., and Yu, K.
\newblock Agentgraph: Towards universal dialogue management with structured deep reinforcement learning.
\newblock \emph{arXiv preprint arXiv:1905.11259}, May 2019.

\bibitem[Costa-jussà et~al.(2022)Costa-jussà, Cross, Çelebi, Elbayad, Heafield, Heffernan, Kalbassi, Lam, Licht, Maillard, Sun, Wang, Wenzek, Youngblood, Akula, Barrault, Gonzalez, Hansanti, Hoffman, Jarrett, Sadagopan, Rowe, Spruit, Tran, Andrews, Ayan, Bhosale, Edunov, Fan, Gao, Goswami, Guzmán, Koehn, Mourachko, Ropers, Saleem, Schwenk, and Wang]{nllbteam2022language}
Costa-jussà, M.~R., Cross, J., Çelebi, O., Elbayad, M., Heafield, K., Heffernan, K., Kalbassi, E., Lam, J., Licht, D., Maillard, J., Sun, A., Wang, S., Wenzek, G., Youngblood, A., Akula, B., Barrault, L., Gonzalez, G.~M., Hansanti, P., Hoffman, J., Jarrett, S., Sadagopan, K.~R., Rowe, D., Spruit, S., Tran, C., Andrews, P., Ayan, N.~F., Bhosale, S., Edunov, S., Fan, A., Gao, C., Goswami, V., Guzmán, F., Koehn, P., Mourachko, A., Ropers, C., Saleem, S., Schwenk, H., and Wang, J.
\newblock No language left behind: Scaling human-centered machine translation.
\newblock \emph{arXiv preprint arXiv:2207.04672}, July 2022.

\bibitem[Greyling(2023)]{greyling-chatgpt-api}
Greyling, C.
\newblock When using the chatgpt api, users will have to manage the context.
\newblock \url{https://cobusgreyling.medium.com/when-using-the-chatgpt-api-users-will-have-to-manage-the-context-ba5869238913}, March 6 2023.

\bibitem[Gu et~al.(2022)Gu, Stefani, Wu, Thomason, and Wang]{Gu_2022}
Gu, J., Stefani, E., Wu, Q., Thomason, J., and Wang, X.
\newblock Vision-and-language navigation: A survey of tasks, methods, and future directions.
\newblock In \emph{Proceedings of the 60th Annual Meeting of the Association for Computational Linguistics (Volume 1: Long Papers)}, Dublin, Ireland, 2022. Association for Computational Linguistics.
\newblock \doi{10.18653/v1/2022.acl-long.524}.

\bibitem[Hao et~al.(2023)Hao, Liu, Wang, and Hu]{hao2023toolkengpt}
Hao, S., Liu, T., Wang, Z., and Hu, Z.
\newblock Toolkengpt: Augmenting frozen language models with massive tools via tool embeddings.
\newblock In \emph{Advances in Neural Information Processing Systems 36}, New Orleans, Louisiana, December 2023.

\bibitem[Hu et~al.(2024)Hu, Huang, Xu, Chen, Xu, Chen, Feng, Wang, Wang, Bao, Sun, and Shan]{Hu2024InferenceWI}
Hu, C., Huang, H., Xu, L., Chen, X., Xu, J., Chen, S., Feng, H., Wang, C., Wang, S., Bao, Y., Sun, N., and Shan, Y.
\newblock Inference without interference: Disaggregate llm inference for mixed downstream workloads.
\newblock \emph{arXiv preprint arXiv:2401.11181}, January 2024.

\bibitem[Hu et~al.(2023)Hu, Lin, Zhang, Yi, and Gao]{hu2023look}
Hu, Y., Lin, F., Zhang, T., Yi, L., and Gao, Y.
\newblock Look before you leap: Unveiling the power of gpt-4v in robotic vision-language planning.
\newblock \emph{arXiv preprint arXiv:2311.17842}, November 2023.

\bibitem[Huang et~al.(2022)Huang, Abbeel, Pathak, and Mordatch]{huang2022language}
Huang, W., Abbeel, P., Pathak, D., and Mordatch, I.
\newblock Language models as zero-shot planners: Extracting actionable knowledge for embodied agents.
\newblock In \emph{Proceedings of 39th International Conference on Machine Learning}, Honolulu, Hawai'i, 2022.

\bibitem[Izacard et~al.(2022)Izacard, Lewis, Lomeli, Hosseini, Petroni, Schick, Dwivedi-Yu, Joulin, Riedel, and Grave]{izacard2022atlas}
Izacard, G., Lewis, P., Lomeli, M., Hosseini, L., Petroni, F., Schick, T., Dwivedi-Yu, J., Joulin, A., Riedel, S., and Grave, E.
\newblock Atlas: Few-shot learning with retrieval augmented language models, 2022.

\bibitem[Khattab et~al.(2024)Khattab, Singhvi, Maheshwari, Zhang, Santhanam, A, Haq, Sharma, Joshi, Moazam, Miller, Zaharia, and Potts]{khattab2024dspy}
Khattab, O., Singhvi, A., Maheshwari, P., Zhang, Z., Santhanam, K., A, S.~V., Haq, S., Sharma, A., Joshi, T.~T., Moazam, H., Miller, H., Zaharia, M., and Potts, C.
\newblock {DSP}y: Compiling declarative language model calls into state-of-the-art pipelines.
\newblock In \emph{The Twelfth International Conference on Learning Representations (ICLR '24)}, Vienna, Austria, 2024.
\newblock URL \url{https://openreview.net/forum?id=sY5N0zY5Od}.

\bibitem[Kwon et~al.(2023)Kwon, Li, Zhuang, Sheng, Zheng, Yu, Gonzalez, Zhang, and Stoica]{vLLM}
Kwon, W., Li, Z., Zhuang, S., Sheng, Y., Zheng, L., Yu, C.~H., Gonzalez, J., Zhang, H., and Stoica, I.
\newblock Efficient memory management for large language model serving with pagedattention.
\newblock In \emph{Proceedings of the 29th Symposium on Operating Systems Principles}, Koblenz, Germany, October 2023.

\bibitem[Li et~al.(2023)Li, Zheng, Zhong, Liu, Sheng, Jin, Huang, Chen, Zhang, Gonzalez, and Stoica]{li2023alpaserve}
Li, Z., Zheng, L., Zhong, Y., Liu, V., Sheng, Y., Jin, X., Huang, Y., Chen, Z., Zhang, H., Gonzalez, J.~E., and Stoica, I.
\newblock {AlpaServe}: Statistical multiplexing with model parallelism for deep learning serving.
\newblock In \emph{17th USENIX Symposium on Operating Systems Design and Implementation (OSDI 23)}, Boston, MA, July 2023.

\bibitem[Lu et~al.(2023)Lu, Peng, Cheng, Galley, Chang, Wu, Zhu, and Gao]{lu2023chameleon}
Lu, P., Peng, B., Cheng, H., Galley, M., Chang, K.-W., Wu, Y.~N., Zhu, S.-C., and Gao, J.
\newblock Chameleon: Plug-and-play compositional reasoning with large language models.
\newblock In \emph{Proceedings of the 37th International Conference on Neural Information Processing Systems (NeurIPS '23)}, New Orleans, Louisiana, December 2023.

\bibitem[Ma et~al.(2015)Ma, Edge, Findlater, and Tan]{ma2015haptic}
Ma, Z., Edge, D., Findlater, L., and Tan, H.~Z.
\newblock Haptic keyclick feedback improves typing speed and reduces typing errors on a flat keyboard.
\newblock In \emph{2015 IEEE World Haptics Conference (WHC)}, Evanston, Illinois, 2015. IEEE.

\bibitem[Meta(2024)]{llama3}
Meta.
\newblock Meta llama 3.
\newblock \url{https://llama.meta.com/llama3/}, 2024.

\bibitem[Mialon et~al.(2023)Mialon, Dessi, Lomeli, Nalmpantis, Pasunuru, Raileanu, Roziere, Schick, Dwivedi-Yu, Celikyilmaz, Grave, LeCun, and Scialom]{mialon2023augmented}
Mialon, G., Dessi, R., Lomeli, M., Nalmpantis, C., Pasunuru, R., Raileanu, R., Roziere, B., Schick, T., Dwivedi-Yu, J., Celikyilmaz, A., Grave, E., LeCun, Y., and Scialom, T.
\newblock Augmented language models: a survey.
\newblock \emph{Transactions on Machine Learning Research (TMLR)}, 2023.
\newblock ISSN 2835-8856.
\newblock URL \url{https://openreview.net/forum?id=jh7wH2AzKK}.
\newblock Survey Certification.

\bibitem[Ng(2024)]{ng-blog}
Ng, A.
\newblock The batch weekly issues 241.
\newblock \url{https://www.deeplearning.ai/the-batch/issue-241/}, March 2024.

\bibitem[NVIDIA(2023)]{nvidia2023fastertransformer}
NVIDIA.
\newblock Fastertransformer.
\newblock \url{https://github.com/NVIDIA/FasterTransformer}, 2023.

\bibitem[{OpenAI}(2023)]{chatgpt-plugin}
{OpenAI}.
\newblock {ChatGPT plugins}.
\newblock \url{https://openai.com/blog/chatgpt-plugins}, March 2023.

\bibitem[OpenTable(2023)]{opentable-chatgpt}
OpenTable.
\newblock New: Chatgpt restaurant recs, powered by opentable.
\newblock \url{https://www.opentable.com/blog/chatgpt/}, March 23 2023.

\bibitem[Parisi et~al.(2022)Parisi, Zhao, and Fiedel]{parisi2022talm}
Parisi, A., Zhao, Y., and Fiedel, N.
\newblock Talm: Tool augmented language models.
\newblock \emph{arXiv preprint arXiv:2205.12255}, May 2022.

\bibitem[Patel et~al.(2024)Patel, Choukse, Zhang, Shah, Goiri, Maleki, and Bianchini]{patel2024splitwise}
Patel, P., Choukse, E., Zhang, C., Shah, A., Goiri, {\'I}., Maleki, S., and Bianchini, R.
\newblock Splitwise: Efficient generative llm inference using phase splitting.
\newblock In \emph{The 53th International Symposium on Computer Architecture (ISCA 2024)}, Buenos Aires, Argentina, June 2024.

\bibitem[Patil et~al.(2023)Patil, Zhang, Wang, and Gonzalez]{patil2023gorilla}
Patil, S.~G., Zhang, T., Wang, X., and Gonzalez, J.~E.
\newblock Gorilla: Large language model connected with massive apis.
\newblock \emph{arXiv preprint arXiv:2305.15334}, 2023.

\bibitem[Qian et~al.(2023)Qian, Han, Fung, Qin, Liu, and Ji]{qian-etal-2023-creator}
Qian, C., Han, C., Fung, Y., Qin, Y., Liu, Z., and Ji, H.
\newblock {CREATOR}: Tool creation for disentangling abstract and concrete reasoning of large language models.
\newblock In Bouamor, H., Pino, J., and Bali, K. (eds.), \emph{Findings of the Association for Computational Linguistics: (EMNLP '23)}, Singapore, December 2023.
\newblock URL \url{https://aclanthology.org/2023.findings-emnlp.462}.

\bibitem[Qin et~al.(2023)Qin, Hu, Lin, Chen, Ding, Cui, Zeng, Huang, Xiao, Han, Fung, Su, Wang, Qian, Tian, Zhu, Liang, Shen, Xu, Zhang, Ye, Li, Tang, Yi, Zhu, Dai, Yan, Cong, Lu, Zhao, Huang, Yan, Han, Sun, Li, Phang, Yang, Wu, Ji, Liu, and Sun]{qin2023tool}
Qin, Y., Hu, S., Lin, Y., Chen, W., Ding, N., Cui, G., Zeng, Z., Huang, Y., Xiao, C., Han, C., Fung, Y.~R., Su, Y., Wang, H., Qian, C., Tian, R., Zhu, K., Liang, S., Shen, X., Xu, B., Zhang, Z., Ye, Y., Li, B., Tang, Z., Yi, J., Zhu, Y., Dai, Z., Yan, L., Cong, X., Lu, Y., Zhao, W., Huang, Y., Yan, J., Han, X., Sun, X., Li, D., Phang, J., Yang, C., Wu, T., Ji, H., Liu, Z., and Sun, M.
\newblock Tool learning with foundation models.
\newblock \emph{arXiv preprint arXiv:2304.08354}, June 2023.

\bibitem[Rombach et~al.(2021)Rombach, Blattmann, Lorenz, Esser, and Ommer]{stable_diffusion}
Rombach, R., Blattmann, A., Lorenz, D., Esser, P., and Ommer, B.
\newblock High-resolution image synthesis with latent diffusion models.
\newblock \emph{arXiv preprint arXiv:2112.10752}, 2021.

\bibitem[Schick et~al.(2023)Schick, Dwivedi-Yu, Dess{\`\i}, Raileanu, Lomeli, Hambro, Zettlemoyer, Cancedda, and Scialom]{schick2023toolformer}
Schick, T., Dwivedi-Yu, J., Dess{\`\i}, R., Raileanu, R., Lomeli, M., Hambro, E., Zettlemoyer, L., Cancedda, N., and Scialom, T.
\newblock Toolformer: Language models can teach themselves to use tools.
\newblock \emph{37th Conference on Neural Information Processing Systems}, 2023.

\bibitem[Shen et~al.(2023)Shen, Song, Tan, Li, Lu, and Zhuang]{shen2023hugginggpt}
Shen, Y., Song, K., Tan, X., Li, D., Lu, W., and Zhuang, Y.
\newblock Hugging{GPT}: Solving {AI} tasks with chat{GPT} and its friends in hugging face.
\newblock In \emph{Proceedings of the 37th International Conference on Neural Information Processing Systems (NeurIPS '23)}, New Orleans, Louisiana, December 2023.

\bibitem[Shoeybi et~al.(2019)Shoeybi, Patwary, Puri, LeGresley, Casper, and Catanzaro]{shoeybi2020megatronlm}
Shoeybi, M., Patwary, M., Puri, R., LeGresley, P., Casper, J., and Catanzaro, B.
\newblock Megatron-lm: Training multi-billion parameter language models using model parallelism.
\newblock \emph{arXiv preprint arXiv:1909.08053}, 2019.

\bibitem[Shridhar et~al.(2021)Shridhar, Yuan, C\^ot\'e, Bisk, Trischler, and Hausknecht]{shridhar2020alfworld}
Shridhar, M., Yuan, X., C\^ot\'e, M.-A., Bisk, Y., Trischler, A., and Hausknecht, M.
\newblock Alfworld: Aligning text and embodied environments for interactive learning.
\newblock In \emph{Proceedings of the International Conference on Learning Representations (ICLR)}, Virtual, 2021.

\bibitem[Srivatsa et~al.(2024)Srivatsa, He, Abhyankar, Li, and Zhang]{srivatsa2024preble}
Srivatsa, V., He, Z., Abhyankar, R., Li, D., and Zhang, Y.
\newblock Preble: Efficient distributed prompt scheduling for llm serving.
\newblock \emph{UCSD CSE Technical Reports}, May 2024.
\newblock URL \url{https://escholarship.org/uc/item/1bm0k1w0}.

\bibitem[Suris et~al.(2023)Suris, Menon, and Vondrick]{surís2023vipergpt}
Suris, D., Menon, S., and Vondrick, C.
\newblock Vipergpt: Visual inference via python execution for reasoning.
\newblock In \emph{2023 IEEE/CVF International Conference on Computer Vision (ICCV '23)}, pp.\  11854--11864, Los Alamitos, CA, USA, October 2023.
\newblock URL \url{https://doi.ieeecomputersociety.org/10.1109/ICCV51070.2023.01092}.

\bibitem[Touvron et~al.(2023)Touvron, Lavril, Izacard, Martinet, Lachaux, Lacroix, Rozi{\`e}re, Goyal, Hambro, Azhar, et~al.]{touvron2023llama}
Touvron, H., Lavril, T., Izacard, G., Martinet, X., Lachaux, M.-A., Lacroix, T., Rozi{\`e}re, B., Goyal, N., Hambro, E., Azhar, F., et~al.
\newblock Llama: Open and efficient foundation language models.
\newblock \emph{arXiv preprint arXiv:2302.13971}, 2023.

\bibitem[Vaidya et~al.(2023)Vaidya, Comly, DeLaere, Patel, and Oh]{tensorRTllm}
Vaidya, N., Comly, N., DeLaere, J., Patel, A., and Oh, F.
\newblock Nvidia tensorrt-llm supercharges large language model inference on nvidia h100 gpus.
\newblock \url{https://developer.nvidia.com/blog/nvidia-tensorrt-llm-supercharges-large-language-model-inference-on-nvidia-h100-gpus/}, 2023.

\bibitem[Wang \& Komatsuzaki(2021)Wang and Komatsuzaki]{gpt-j-6b}
Wang, B. and Komatsuzaki, A.
\newblock {GPT-J-6B: A 6 Billion Parameter Autoregressive Language Model}.
\newblock \url{https://github.com/kingoflolz/mesh-transformer-jax}, May 2021.

\bibitem[Wang et~al.(2023)Wang, Ma, Feng, Zhang, ran Yang, Zhang, Chen, Tang, Chen, Lin, Zhao, Wei, and rong Wen]{Wang2023ASO}
Wang, L., Ma, C., Feng, X., Zhang, Z., ran Yang, H., Zhang, J., Chen, Z.-Y., Tang, J., Chen, X., Lin, Y., Zhao, W.~X., Wei, Z., and rong Wen, J.
\newblock A survey on large language model based autonomous agents.
\newblock \emph{arXiv preprint arXiv:2308.11432}, August 2023.
\newblock URL \url{https://api.semanticscholar.org/CorpusID:261064713}.

\bibitem[Wei et~al.(2023)Wei, Wang, Schuurmans, Bosma, Ichter, Xia, Chi, Le, and Zhou]{wei2022chain}
Wei, J., Wang, X., Schuurmans, D., Bosma, M., Ichter, B., Xia, F., Chi, E.~H., Le, Q.~V., and Zhou, D.
\newblock Chain-of-thought prompting elicits reasoning in large language models.
\newblock In \emph{Proceedings of the 36th International Conference on Neural Information Processing Systems}, New Orleans, Louisiana, 2023.

\bibitem[Wolfram(2023)]{wolfram-chatgpt}
Wolfram, S.
\newblock Chatgpt gets its `wolfram superpowers'!
\newblock \url{{https://writings.stephenwolfram.com/2023/03/chatgpt-gets-its-wolfram-superpowers/}}, March 2023.

\bibitem[Wu et~al.(2023)Wu, Zhong, Zhang, Huang, Liu, and Jin]{wu2023fast}
Wu, B., Zhong, Y., Zhang, Z., Huang, G., Liu, X., and Jin, X.
\newblock Fast distributed inference serving for large language models.
\newblock \emph{arXiv preprint arXiv:2305.05920}, May 2023.

\bibitem[Yang et~al.(2018)Yang, Qi, Zhang, Bengio, Cohen, Salakhutdinov, and Manning]{yang2018hotpotqa}
Yang, Z., Qi, P., Zhang, S., Bengio, Y., Cohen, W.~W., Salakhutdinov, R., and Manning, C.~D.
\newblock Hotpotqa: A dataset for diverse, explainable multi-hop question answering.
\newblock \emph{arXiv preprint arXiv:1809.09600}, 2018.

\bibitem[Yao et~al.(2023)Yao, Zhao, Yu, Du, Shafran, Narasimhan, and Cao]{yao2022react}
Yao, S., Zhao, J., Yu, D., Du, N., Shafran, I., Narasimhan, K.~R., and Cao, Y.
\newblock React: Synergizing reasoning and acting in language models.
\newblock In \emph{The Eleventh International Conference on Learning Representations}, Kigali, Rwanda, May 2023.

\bibitem[Yu et~al.(2022)Yu, Jeong, Kim, Kim, and Chun]{Orca}
Yu, G.-I., Jeong, J.~S., Kim, G.-W., Kim, S., and Chun, B.-G.
\newblock {Orca: A Distributed Serving System for {Transformer-Based} Generative Models}.
\newblock In \emph{16th USENIX Symposium on Operating Systems Design and Implementation (OSDI '22)}, Carlsbad, CA, July 2022.

\bibitem[Zhang et~al.(2023)Zhang, Zhang, Li, and Smola]{zhang2023automatic}
Zhang, Z., Zhang, A., Li, M., and Smola, A.
\newblock Automatic chain of thought prompting in large language models.
\newblock In \emph{The Eleventh International Conference on Learning Representations}, Kigali, Rwanda, May 2023.

\bibitem[Zheng et~al.(2023{\natexlab{a}})Zheng, Chiang, Sheng, Zhuang, Wu, Zhuang, Lin, Li, Li, Xing, Zhang, Gonzalez, and Stoica]{vicuna_share_gpt}
Zheng, L., Chiang, W.-L., Sheng, Y., Zhuang, S., Wu, Z., Zhuang, Y., Lin, Z., Li, Z., Li, D., Xing, E., Zhang, H., Gonzalez, J.~E., and Stoica, I.
\newblock Judging {LLM}-as-a-judge with {MT}-bench and chatbot arena.
\newblock In \emph{Thirty-seventh Conference on Neural Information Processing Systems Datasets and Benchmarks Track}, New Orleans, Louisiana, December 2023{\natexlab{a}}.

\bibitem[Zheng et~al.(2023{\natexlab{b}})Zheng, Yin, Xie, Huang, Sun, Yu, Cao, Kozyrakis, Stoica, Gonzalez, Barrett, and Sheng]{zheng2023efficiently}
Zheng, L., Yin, L., Xie, Z., Huang, J., Sun, C., Yu, C.~H., Cao, S., Kozyrakis, C., Stoica, I., Gonzalez, J.~E., Barrett, C., and Sheng, Y.
\newblock Efficiently programming large language models using sglang.
\newblock \emph{arXiv preprint arXiv:2312.07104}, December 2023{\natexlab{b}}.

\bibitem[Zhong et~al.(2024)Zhong, Liu, Chen, Hu, Zhu, Liu, Jin, and Zhang]{zhong2024distserve}
Zhong, Y., Liu, S., Chen, J., Hu, J., Zhu, Y., Liu, X., Jin, X., and Zhang, H.
\newblock Distllm: Disaggregating prefill and decoding for goodput-optimized large language model serving.
\newblock In \emph{Proceedings of the 18th USENIX Symposium on Operating Systems Design and Implementation (OSDI '24)}, Santa Clara, CA, July 2024.

\end{thebibliography}



\newpage
\appendix
\onecolumn

{
\begin{figure*}[t]
\begin{center}
\centerline{\includegraphics[width=\textwidth]{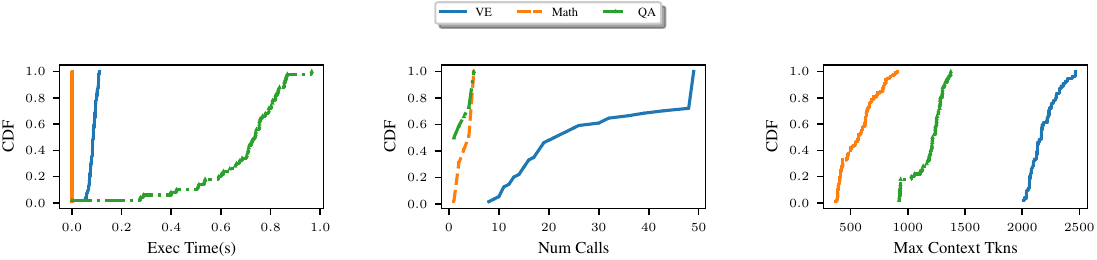}}
\mycaption{fig-api-study-1}{CDF Results of Short APIs.}
{
Each line plots the CDF distribution of all the calls of one API type. 
}
\end{center}
\end{figure*}
}


{
\begin{figure*}[t]
\begin{center}
\centerline{\includegraphics[width=\textwidth]{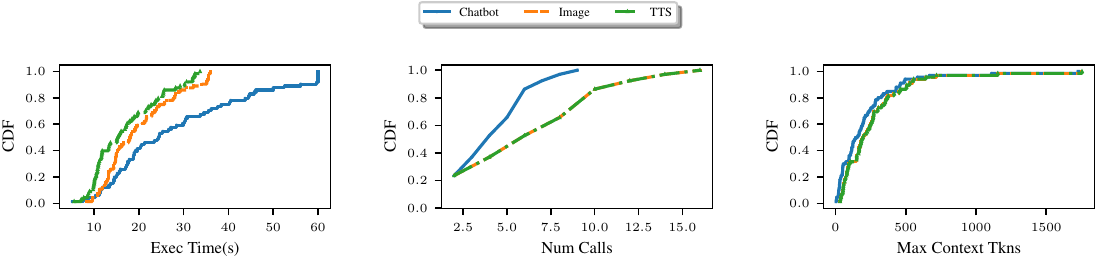}}
\mycaption{fig-api-study-2}{CDF Results of Long APIs.}
{
Each line plots the CDF distribution of all the calls of one API type. 
}
\end{center}
\end{figure*}
}

\section*{API Properties and Dataset Generation}

We analyze the properties of different datasets to evaluate the effectiveness of our inference system. We consider datasets from arithmetic, a knowledge-based question and answering system, a multi-step chat bot, and an embodied virtual environment. The dataset demonstrates different execution time, frequency of API calls, and context lengths. We augment the datasets with relevant APIs from calculator, wikipedia, and a virtual environment gym.

We measure the API execution time, the number of API calls, the number of returned tokens of an API call, and the context length when an API is called.
Figures~\ref{fig-api-study-1} and \ref{fig-api-study-2} present the detailed CDF results the four metrics for short-running APIs (math, QA, VE) and long-running APIs (chatbot, image, TTS).
Below, we briefly introduce each API type and discuss their properties.

We use the ReAct framework to prompt the LLM to call a Wikipedia endpoint with different questions. The responses from the Wikipedia endpoint are postprocessed. We use the live Wikipedia API endpoint for our benchmark and For each call, we retrieve a summary of the relevant Wikipedia page, which we post-process by limiting the size of the responses to fit within the maximum model sequence length.

To evaluate VE, we prompt GPT-4 LLM with an initial virtual environment and use the ReAct event loop to repeatedly interact with the environment. For our evaluation, we truncate to fit within the context length.

We prompt GPT-4 to generate complex Stable Diffusion prompts at a certain size. Each prompt will trigger an API call of the Stable Diffusion model~\cite{stable_diffusion} to generate an image. Each round with the LLM involves a user prompt with normal decoding and then a diffusion call. The number of calls total is modeled around the chat dataset.
We estimate the number of image generation rounds (\ie, number of API calls) to be the same as the number of chatting rounds in Chatbot. 
As there is no standard way of returning tokens from an image generation, we use a short, constant-length sentence describing the generated image as returned tokens. 
For API execution time, we measure the actual Stable Diffusion call and estimate the human response time in the same as as Chatbot; the sum of them at each round is shown in Table~\ref{tbl-api-study} and Figure~\ref{fig-api-study-2}. For the context and return length, we provide a mixture of ShareGPT dataset~\cite{vicuna_share_gpt} and the real prompts. 

We use a methodology similar to image generation to understand TTS-augmented LLMs. 

{
\begin{figure*}[th]
\begin{center}
\centerline{\includegraphics[width=0.6\textwidth]{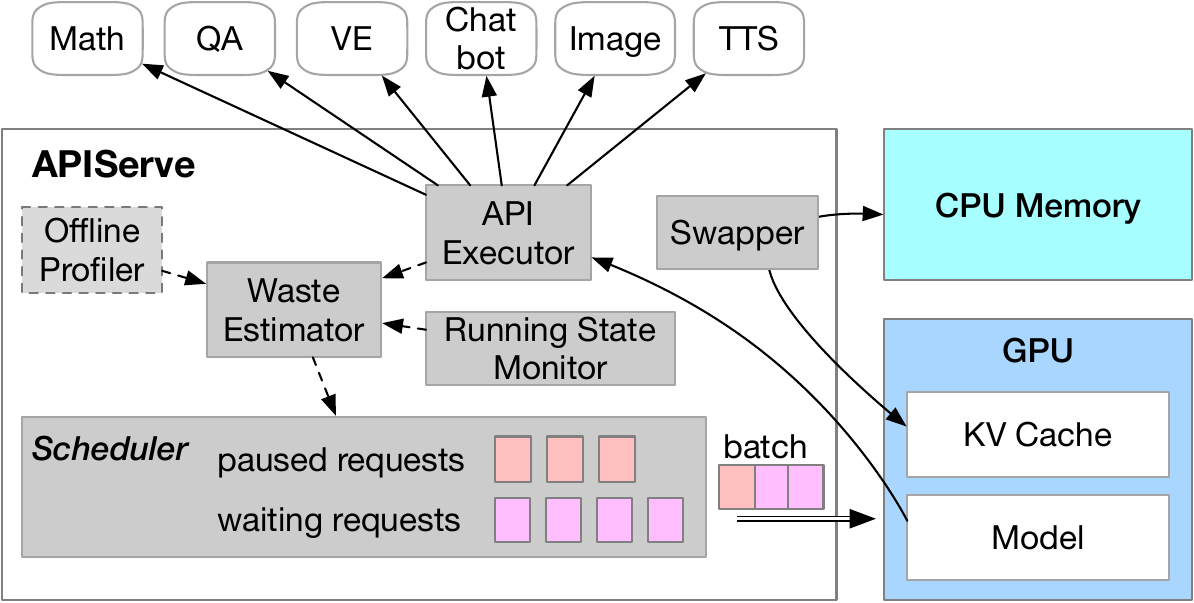}}
\mycaption{fig-arch}{Overall \sys\ Architecture.}
{
}
\end{center}
\end{figure*}
}

\section*{Implementation Details}
\label{sec:impl}


We implement \trick\ on top of vLLM and build {\em \sys}, an inference system designed for API-augmented LLMs. As shown in Figure~\ref{fig-arch}, \sys\ includes an API executor that performs different API calls, an iterative-level scheduler that forms a batch of waiting requests and (partial) paused requests, a swap manager that facilitates GPU/CPU memory swapping, a running status monitor that monitors the resource usage and size of running requests, a waste estimator that calculates GPU memory waste for each API-paused request, and an offline profiler that collects basic systems metrics before starting serving.

At the end of each iteration, the scheduler gathers all requests that trigger API calls, adds them to the paused-request queue, swaps as much context (computed keys and values) of paused requests as allowed to CPU memory, determines whether to preserve or discard the remaining context of each paused request based on the GPU memory waste amount calculated by the waste estimator, and calls the actual API. When an API call finishes, \sys\ determines how many swapped-out or discarded tokens to swap in or recompute in the next iteration. It also determines how many waiting requests (non-API paused requests) to schedule in the next iteration. When all the context of a paused request has been restored, \sys\ adds the tokens returned by the API call to the resumed request.


\end{document}